\documentclass[dvipsnames,pbalance,format=sigconf,anonymous=false,review=false, nonacm=true]{acmart}

\usepackage[inline]{enumitem}
\usepackage{hyperref}
\usepackage[nameinlink]{cleveref}
\usepackage{glossaries}
\usepackage[dvipsnames]{xcolor}

\usepackage{graphicx} 
\usepackage{algorithm}
\usepackage{algorithmicx}
\usepackage{algpseudocode}

\usepackage[most]{tcolorbox}
\usepackage{placeins}

\tcbset {
  base/.style={
    arc=0mm, 
    bottomtitle=0.5mm,
    boxrule=0mm,
    colbacktitle=black!10!white, 
    coltitle=black, 
    fonttitle=\bfseries, 
    left=2.5mm,
    leftrule=1mm,
    right=3.5mm,
    title={#1},
    toptitle=0.75mm, 
    breakable
  }
}

\definecolor{brandblue}{rgb}{0.34, 0.7, 1}
\newtcolorbox{mainbox}[1]{
  colframe=brandblue, 
  base={#1}
}

\newtcolorbox{subbox}[1]{
  colframe=black!30!white,
  base={#1}
}

\hypersetup{
  colorlinks=true, 
  urlcolor=black, 
  linkcolor=black, 
  citecolor=black 
}

\newacronym{aaai}{AAAI}{Association for the Advancement of Artificial Intelligence}
\newacronym{acm}{ACM}{Association for Computing Machinery}
\newacronym{ai}{AI}{Artificial Intelligence}
\newacronym{bbsr}{BBSR}{Benchmarking, Benchmarks, Software and Reproducibility}
\newacronym{ec}{EC}{Evolutionary Computation}
\newacronym{gecco}{GECCO}{Genetic and Evolutionary Computation Conference}
\newacronym{llm}{LLM}{Large Language Model}
\newacronym{ml}{ML}{Machine Learning}
\newacronym{neurips}{NeurIPS}{Neural Information Processing Systems}
\newacronym{aad}{AAD}{Automated Algorithm Design}
\newacronym{ast}{AST}{Abstract Syntax Tree}
\newacronym{cf}{CF}{Code Features}

\setcopyright{acmlicensed}
\acmDOI{10.1145/nnnnnnn.nnnnnnn} 
\acmISBN{978-x-xxxx-xxxx-x/YY/MM} 
\acmConference[GECCO '26]{The Genetic and Evolutionary Computation Conference 2026}{July 13--17, 2026}{San José, Costa Rica}
\acmYear{2026}
\copyrightyear{2026}

\begin{document}

\title{LLaMEA-SAGE: Guiding Automated Algorithm Design with Structural Feedback from Explainable AI}


\author{Niki van Stein}
\email{n.van.stein@liacs.leidenuniv.nl}
\orcid{0000-0002-0013-7969}
\affiliation{%
  \institution{LIACS, Leiden University}
  \country{Netherlands}
}

\author{Anna V. Kononova}
\email{a.kononova@liacs.leidenuniv.nl}
\orcid{0000-0002-4138-7024}
\affiliation{%
  \institution{LIACS, Leiden University}
  \country{Netherlands}
}

\author{Lars Kotthoff}
\email{lk223@st-andrews.ac.uk}
\orcid{0000-0003-4635-6873}
\affiliation{%
  \institution{University of St Andrews}
  \country{United Kingdom}
}
\author{Thomas B{\"a}ck}
\email{t.h.w.baeck@liacs.leidenuniv.nl}
\orcid{0000-0001-6768-1478}
\affiliation{%
  \institution{LIACS, Leiden University}
  \country{Netherlands}
}

\renewcommand{\shortauthors}{van Stein et al.}

\begin{abstract}
Large language models have enabled automated algorithm design (AAD) by generating optimization algorithms directly from natural-language prompts. While evolutionary frameworks such as LLaMEA demonstrate strong exploratory capabilities across the algorithm design space, their search dynamics are entirely driven by fitness feedback, leaving substantial information about the generated code unused. We propose a mechanism for guiding AAD using feedback constructed from graph-theoretic and complexity features extracted from the abstract syntax trees of the generated algorithms, based on a surrogate model learned over an archive of evaluated solutions. Using explainable AI techniques, we identify features that substantially affect performance and translate them into natural-language mutation instructions that steer subsequent LLM-based code generation without restricting expressivity.

We propose LLaMEA-SAGE, which integrates this feature-driven guidance into LLaMEA, and evaluate it across several benchmarks. We show that the proposed structured guidance achieves the same performance faster than vanilla LLaMEA in a small controlled experiment. In a larger-scale experiment using the MA-BBOB suite from the GECCO-MA-BBOB competition, our guided approach achieves superior performance compared to state-of-the-art AAD methods. These results demonstrate that signals derived from code can effectively bias LLM-driven algorithm evolution, bridging the gap between code structure and human-understandable performance feedback in automated algorithm design.
\end{abstract}

\ccsdesc[500]{Computing methodologies~Heuristic function construction}
\ccsdesc[300]{Computing methodologies~Continuous space search}
\ccsdesc[300]{Computing methodologies~Natural language generation}

\keywords{Automated Algorithm Design, Large Language Models, Evolutionary Computation, Structural code analysis, Explainable AI–guided optimization}

\maketitle

\glsresetall

\begin{figure*}[!tb]
    \centering
    \includegraphics[width=0.95\linewidth]{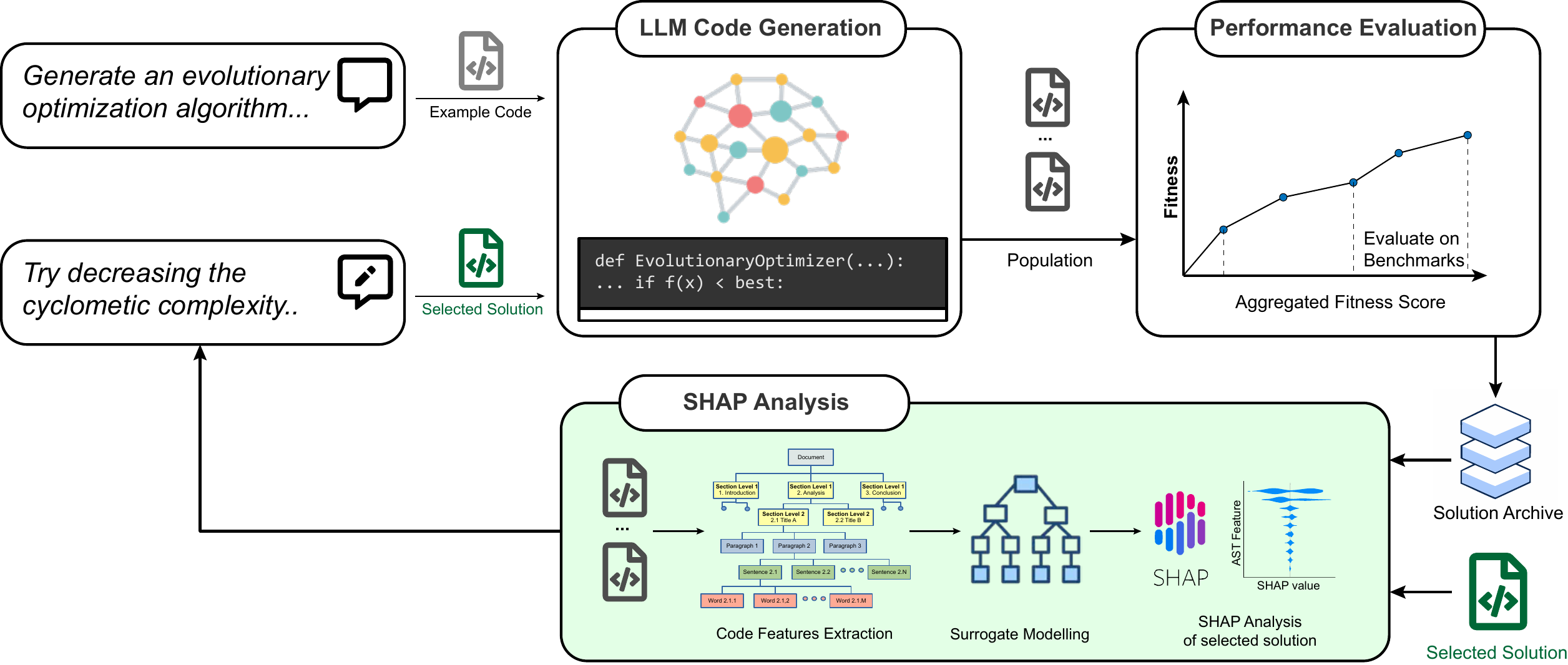}
    \caption{Overview of the proposed method \textit{LLaMEA-SAGE}.}
    \label{fig:method}
\end{figure*}

\section{Introduction}\label{sec:introduction}
Designing effective and efficient optimization algorithms remains a core challenge in \gls{ec}~\cite{Baeck2023}. While decades of research have produced highly successful hand-crafted methods, such as CMA-ES~\cite{hansen2003reducing}, their design typically relies on substantial human expertise, iterative trial-and-error and implicit domain knowledge. This makes systematic algorithm innovation slow, difficult to replicate and hard to scale across problem classes.

Recent advances in large language models (LLMs) have opened up a new paradigm for \gls{aad}, where algorithm discovery itself is framed as a search problem over executable programs. A growing body of work demonstrates that LLMs can generate, tune hyperparameters, modify and refine heuristics or algorithms when embedded in structured search procedures \cite{stein2025llamea,fei2024eoh,zhengmonte,LHNS,stein2025llameahpo}. These approaches show that LLMs can contribute to algorithmic innovation rather than merely code synthesis.

Despite this progress, several important shortcomings remain. First, many LLM-based heuristic design methods rely on fixed algorithmic templates, limiting the scope of discoverable algorithms and biasing the search toward known structures. Second, search strategies often struggle to balance exploration and exploitation in large, heterogeneous code spaces, leading to either premature stagnation or inefficient use of expensive LLM queries. Third, while promising results have been reported, there is still limited understanding of how to systematically improve LLM-generated algorithms after the initial generation of the code and how such improvements compare to state-of-the-art LLM-based heuristic design methods under controlled benchmarking~\cite{benchmarkingArxiv2020}.

In this paper, we address some of these limitations by extending the LLaMEA framework with a structured feedback mechanism. During the algorithm discovery process, it learns which code features of the discovered algorithms are correlated with performance and may lead to better results. We focus on scalable, template-free evolution of full black-box optimization algorithms, while explicitly positioning our approach relative to recent state-of-the-art LLM-based heuristic design methods, including MCTS-AHD~\cite{zhengmonte} and LHNS~\cite{LHNS}. 

Our main contributions are as follows:
\begin{itemize}
    \item We propose \textit{LLaMEA-SAGE} which guides the algorithm discovery process using explainable AI analysis of code features, providing structured feedback to the LLM.
    \item We conduct a systematic experimental comparison against state-of-the-art LLM-based heuristic design methods, including MCTS-AHD and LHNS, using them as strong baselines alongside the vanilla LLaMEA.
    \item We empirically demonstrate that the proposed approach improves algorithm quality and robustness on black-box optimization benchmarks and provide an extensive analysis of the \gls{aad} runs.
\end{itemize}
All code and results are available in our repository\footnote{\url{https://anonymous.4open.science/r/LLaMEA-SAGE/README.md}}.

\section{Related Work}\label{sec:rel-work}
One of the earliest frameworks in the \gls{aad} area is \emph{Evolution of Heuristics} (EoH)~\cite{fei2024eoh}. EoH integrates LLMs into an evolutionary computation loop, where candidate heuristics are represented by both natural-language descriptions and corresponding code fragments. New heuristics are generated through LLM-driven mutation and crossover operations, guided by performance-based selection. EoH has demonstrated strong performance on a range of combinatorial optimization problems and established LLM-based heuristic evolution as a competitive alternative to hand-crafted designs and other program search approaches such as FunSearch \cite{FunSearch2024}.

Several follow-up works have explored alternative search strategies to better cope with the size and complexity of the heuristic search space \cite{chauhan2025evolutionary}. \emph{MCTS-AHD} replaces the population-based EC loop with a Monte Carlo Tree Search (MCTS) formulation, organizing all generated heuristics in a tree structure and using MCTS to balance exploration and exploitation~\cite{zhengmonte}. By allowing temporarily underperforming heuristics to be revisited and refined, MCTS-AHD mitigates premature stagnation. 

Another recent state-of-the-art method in LLM-based AAD is \emph{LLM-driven Heuristic Neighborhood Search} (LHNS)~\cite{LHNS}, which abandons population-based evolution in favor of a single-solution neighborhood search paradigm, iteratively applying ruin-and-recreate operations to partially destroy and reconstruct heuristic code using an LLM. 

In contrast to these heuristic-centric approaches, \emph{LLaMEA}~\cite{stein2025llamea} targets automated discovery of \textit{complete} optimization algorithms rather than individual heuristic components. LLaMEA embeds LLMs within the framework of evolutionary strategies, such as $(1,1)$, $(1{+}1)$ and $(\mu{+}\lambda)$ schemes, and directly optimizes executable algorithm implementations using runtime performance on selected benchmarks as feedback. Unlike EoH-style methods, LLaMEA does not rely on predefined heuristic slots or algorithmic templates, enabling the generation and optimization of larger and more flexible code bases. On continuous black-box optimization benchmarks, LLaMEA has been shown to produce algorithms that are competitive with, and in some cases outperform, established methods such as CMA-ES \cite{hansenCMAEvolutionStrategy2023a} and differential evolution \cite{storn1997differential}. Subsequent extensions, such as LLaMEA-HPO \cite{stein2025llameahpo}, further decouple algorithm structure discovery from hyperparameter tuning, improving sample efficiency.

\paragraph{Positioning}
The methods above illustrate complementary trade-offs in LLM-based algorithm design. EoH, MCTS-AHD and LHNS focus on \textit{structured} heuristic search within predefined frameworks and have achieved strong results on \textit{combinatorial} optimization problems. LLaMEA, in contrast, emphasizes end-to-end discovery of \textit{full} algorithms in \textit{continuous} black-box optimization. In this work, we build on the open-source and modular framework of LLaMEA as a flexible foundation for black-box optimization algorithm design, while using vanilla LLaMEA, MCTS-AHD and LHNS as state-of-the-art baselines for comparison. This allows us to assess progress incrementally, relative to leading LLM-based heuristic design approaches while focusing on scalable, template-free algorithm discovery.

\paragraph{Code features for algorithm design}
Static code analysis is a well-established technique in software engineering and has been important in assessing the maintainability and quality of source code \cite{nunez2017source}.
The features classically used for static code analysis have also shown their use for better selecting and configuring algorithms~\cite{pulatov2022opening}. Pulatov et al.~\cite{pulatov2022opening} introduce an approach to improving algorithm selection by analysing structural features of algorithms, moving beyond traditional ``black-box'' methods that rely solely on performance observations. The authors demonstrate that incorporating algorithm features into the selection process can lead to performance improvements. We further this idea by integrating code-based features into automated algorithm design with the goal of making the complex and open-ended evolution of code more efficient and using the information gathered during the evolutionary process as effectively as possible.

\section{LLaMEA-SAGE: Structural Code-feature Guided Algorithm Evolution}
\label{sec:methodology}

We propose a \emph{code-features-driven mutation guidance mechanism} for automated algorithm design that augments LLM-based evolutionary search with structural signals extracted from program code (see Algorithm \ref{alg:ast-guided-aad}). The method is implemented in the open-source LLaMEA framework \cite{stein2025llamea} and introduces an archive-based surrogate model that learns relations between algorithmic code structure and optimization performance. These relations are analysed and turned into lightweight natural language descriptions, which augment the LLM prompt to guide subsequent modifications.

\begin{algorithm}[t]
\caption{LLaMEA-SAGE}
\label{alg:ast-guided-aad}
\begin{algorithmic}[1]
\Require Evaluation function $f(\cdot)$, LLM $\mathcal{M}$, mutation prompts $\mathcal{P}$,
Feature computation function \texttt{cf}$(\cdot$),
Population size $\mu$, offspring size $\lambda$, evaluation budget $B$
\State Initialize population $P$ using $\mathcal{M}$
\State Evaluate, compute code features (cf) and store $(s, f(s), \texttt{cf}(s))$ for all $s \in P$ in archive $\mathcal{A}$
\While{$|\mathcal{A}| < B$}
    \State Train surrogate model $\hat{f}$ on code features from $\mathcal{A}$
    \State $O \gets \emptyset$
    \For{$i = 1$ to $\lambda$}
        \State Select parent $p \in P$
        \State Sample mutation prompt $q \in \mathcal{P}$
        \State Augment $q$ to increase/decrease code feature value with highest impact on performance given $p$
        \State Generate offspring $s'$ using LLM $\mathcal{M}$, $p$ and prompt $q$
        \State Evaluate $f(s')$ and extract code features
        \State Add $(s', f(s'), \texttt{cf}(s'))$ to archive $\mathcal{A}$
        \State Add $s'$ to offspring set $O$
    \EndFor
    \State Select next population $P$ from $P \cup O$ (elitist selection)
\EndWhile
\State \Return best solution in $\mathcal{A}$
\end{algorithmic}
\end{algorithm}

\subsection{Algorithm Representation via Abstract Syntax Trees}
Each candidate algorithm is represented as executable Python code. The correctness of the code is assessed during the evaluation. To enable learning over structural properties of algorithms, we extract a rich set of features from the code’s \gls{ast}. Given a solution code  $c$, we parse it into an \gls{ast} and construct a directed graph
\[
G_c = (V, E),
\]
where nodes correspond to code-feature elements and edges encode parent--child relations in the syntax tree \cite{neamtiu2005understanding}.

From $G_c$, we compute graph-theoretic statistics capturing size, depth, connectivity and heterogeneity, including:
\begin{itemize}
    \item node and edge counts,
    \item degree statistics (mean, variance, entropy),
    \item tree depth statistics (min, mean, max),
    \item clustering coefficients and assortativity,
    \item path-based metrics such as diameter and average shortest path.
\end{itemize}

In addition, we extract code-level complexity indicators using static analysis, including cyclomatic complexity, function token counts and parameter counts aggregated across functions. The final feature vector for a solution $s$ is
\[
\mathbf{x}_s \in \mathbb{R}^d,
\]
where $d$ denotes the total number of AST and complexity features. The full code features extraction pipeline is deterministic and lightweight, enabling reuse during evolution without re-calculating code features for unchanged solutions per iteration.  

\subsection{Archive-Based Surrogate Modelling}

During evolution, all evaluated solutions including extracted code features (cf) are stored in an archive $\mathcal{A} = \{(s_i, f_i, \texttt{cf}_i)\}_{i=1}^N$, where $f_i$ denotes the observed fitness and $\texttt{cf}_i$ the features derived from the \gls{ast} and code of $i$.

Using the archive, we train a gradient-boosted regression tree model
\[
\hat{f}(cf) \approx f,
\]
using XGBoost \cite{chen2015xgboost} with squared-error loss to approximate the code-features--fitness relation function $f$ given code features $cf$. Training is triggered once a minimum archive size is reached to avoid degenerate models. This surrogate captures non-linear relations between code structure and empirical performance.

\subsection{Feature Importance via SHAP}

To derive actionable mutation guidance, we apply SHAP~\cite{lundberg2020local2global} to the trained surrogate model. For each feature $j$, SHAP provides an attribution value $\phi_j$ quantifying its contribution to the predicted fitness:
\[
\hat{f}(\mathbf{x}) = \phi_0 + \sum_{j=1}^{d} \phi_j.
\]


The feature with the largest absolute attribution is selected and the sign of its SHAP value determines whether the feature value should be \emph{increased} or \emph{decreased}.

\subsection{Guided Mutation Prompting}

The resulting guidance is translated into a natural-language instruction that augments the standard mutation prompt provided to the LLM. Formally, if feature $k$ is selected with action $a \in \{\text{increase}, \text{decrease}\}$, the mutation operator is extended with:
\begin{mainbox}{Guided Mutation Prompt}
\emph{``Based on archive analysis, try to \texttt{$<$a$>$} the \texttt{$<$k$>$} of the solution.''}
\end{mainbox}

This guidance does not constrain the LLM syntactically but biases the generated mutation toward structural regions that are empirically correlated with improved performance. We do not claim that these found correlations are causally correct, but even imperfect, noisy correlations are sufficient to bias exploration beneficially in early \gls{aad} stages as we will show in the results section.
The evolutionary loop remains the same as LLaMEA otherwise, preserving the exploratory nature of the LLM-driven algorithm synthesis.

Overall, the method closes the loop between \emph{code structure}, \emph{observed optimization performance} and \emph{future algorithm generation}, yielding a data-driven form of inductive bias for automated algorithm design.  

Note that while we evaluated our structured guidance approach with LLaMEA, the general approach is not specific to any particular framework and could also be used with EoH, MCTS-AHD or LHNS.

\section{Experimental setup}\label{sec:exp}
We evaluate the proposed code-features driven guidance, LLaMEA-SAGE, in two AAD experiments designed to isolate its effect and assess scalability to established benchmarking scenarios. All experiments are done using the IOH-BLADE 
\cite{van2025blade} benchmarking toolbox; raw results, code and post-processing notebooks are available in our repository\footnote{\url{https://anonymous.4open.science/r/LLaMEA-SAGE/README.md}}. Final discovered algorithms are additionally validated on unseen instances using different dimensionalities and evaluation budgets to assess generalizability. In addition we assess their performance against SOTA baselines. 

\subsection{Performance Metric}
For all experiments, we use an anytime performance metric, AOCC, to access the quality of generated black-box optimizers on selected benchmarks and as a fitness score feedback to the LLM. Eq.~\ref{eq:aocc} gives the definition of AOCC:
\begin{equation}
\resizebox{0.9\linewidth}{!}{
    $AOCC(y_{a,f}) = \frac{1}{B}\sum_{i=1}^{B} \left ( 1 - \frac{\min(\max((y_i),lb),ub) - lb}{ub-lb} \right )$
}
\label{eq:aocc}
\end{equation}
where $y_{a,f}$ is a series of log-scaled current-best fitness value of $f$ during the optimization algorithm run $a$, $B$ is the evaluation budget, $y_i$ is the $i$-th element of $y_{a,f}$, $ub$ is the upper bound of $f$ and $lb$ is the lower bound of $f$.

\subsection{Experiment~1: Vanilla LLaMEA vs.\ Feature-Guided LLaMEA-SAGE}\label{sec:experiment1}

\paragraph{Objective}
The first experiment isolates the effect of our proposed feature-guided mutation by comparing a vanilla LLaMEA-based evolutionary strategy (LLaMEA) against an identical strategy augmented with code-features-based guidance (LLaMEA-SAGE).

\paragraph{Benchmark}
We use a subset of the SBOX-COST benchmark suite \cite{vermettenSBOXcost}, focusing on multiple noiseless single objective box-constrained optimization problems. 
Functions $f \in \{1,\dots,5\}$ are used\\
(all separable functions), each evaluated across $5$ instances. \\ Dimensionality is fixed to $d=10$.

\paragraph{Methods}
Two methods are compared:
\begin{itemize}
    \item \textbf{LLaMEA}: Standard LLaMEA with random parent selection and standard mutation prompts.
    \item \textbf{LLaMEA-SAGE}: Identical configuration but now with feature guided mutation enabled.
\end{itemize}

Both methods use $\mu=8$ parents, $\lambda=8$ offspring, elitist selection and an identical LLM backend (GPT-5-mini). The total evolutionary budget is fixed to $200$ evaluations. The two mutation prompts used were \textit{``Generate a new algorithm that is different from the algorithms you have tried before.''} and \textit{``Refine the strategy of the selected solution to improve it.''} For the remainder of the paper we refer to these prompts as \textbf{random new} and \textbf{refine} respectively. We use the initial population size (in this case $8$) as the minimum archive size before training a surrogate model.

\paragraph{Evaluation Protocol}
Each method (LLaMEA and LLaMEA-SAGE) is repeated for $5$ independent runs with different random seeds. Performance is measured using current-best fitness (AOCC) over each run and aggregated in the figures using mean and  95\% confidence intervals across runs.

\subsection{Experiment~2: Comparison against SOTA}\label{sec:experiment2}
\paragraph{Objective}
The second experiment evaluates whether feature-guided mutation improves performance in a multi-algorithm, multi-function setting comparing to the latest state-of-the-art methods in automated algorithm design.

\paragraph{Benchmark}
We use the Many Affine BBOB benchmark (MA-BBOB) \cite{vermetten2023ma} as this is also an official GECCO competition and contains a large variety of different functions. In this setup, we evaluate generated algorithms across multiple affine BBOB functions. Training is performed on $10$ instances, with dimensionality $d=10$ and a budget of $5000 \cdot d$ evaluations per algorithm.

\paragraph{Methods}
This experiment positions the proposed method against established automated algorithm design baselines.
\begin{itemize}
    \item \textbf{LLaMEA-SAGE}: LLaMEA with code feature--driven mutation guidance. $4$ parents and $16$ offspring, elitism enabled.
    \item \textbf{LLaMEA}: vanilla LLaMEA without feature guidance. $4$ parents and $16$ offspring, elitism enabled.
    \item \textbf{MCTS-based algorithm synthesis} using recommended parameters from \cite{zhengmonte}.
    \item \textbf{LHNS} (Large Neighborhood Search over code), using recommended parameters from \cite{LHNS}.
\end{itemize}

Each method is executed for $5$ independent runs with seeds $\{1,\dots,5\}$, all using the same LLM backend (GPT-5-mini) and evaluation budget. Each generated algorithm as a 1-hour time-limit to complete (to prevent non-terminating algorithms).
All methods are evaluated on MA-BBOB with identical training instances, dimensionality, evaluation budgets and random seeds. Each method is allowed the same number of algorithm evaluations (200), ensuring a fair comparison.

Each method uses the exact same problem formulation prompt to start with (See Appendix A).
Performance is evaluated using aggregated anytime performance (AOCC) over MA-BBOB tasks, following the standard IOHProfiler logging and aggregation protocol \cite{IOHprofiler}.

\section{Results}\label{sec:results}

\subsection{Proof of concept experiment}
\begin{figure}[!tb]
    \centering
    \includegraphics[width=1.0\linewidth,trim=3mm 4mm 4mm 9mm,clip]{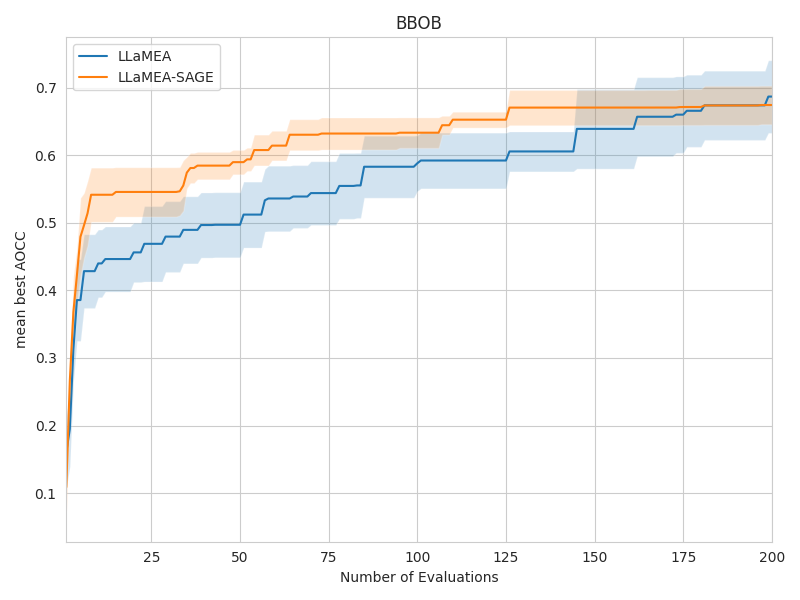}
    \caption{Average best-so-far fitness (AOCC) over time of the baseline LLaMEA and the proposed code-feature guided approach (LLaMEA-SAGE) on the SBOX-COST suite. Aggregated over $5$ independent runs, the 95\% confidence interval is shown as shaded area.}
    \label{fig:SBOX}
\end{figure}

\begin{figure}[!tb]
    \centering
    \includegraphics[width=1.0\linewidth,trim=3mm 0mm 4mm 0mm,clip]{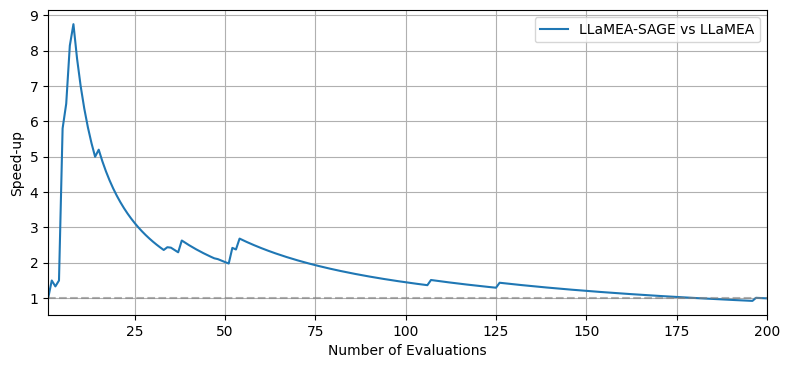}
    \caption{Average speed-up of LLaMEA-SAGE versus vanilla LLaMEA. The speed-up is how much faster than LLaMEA the proposed method is, i.e.\ if LLaMEA-SAGE reaches fitness value $f_i$ in $m$ evaluations and LLaMEA reaches that same fitness value in $n$ evaluations, the speed-up is $n/m$.}
    \label{fig:SBOX-speedup}
\end{figure}


In Figure \ref{fig:SBOX}, we can observe the aggregated performance of the proposed method versus vanilla LLaMEA (see Section~\ref{sec:experiment1} for the setup description). We can see that, especially at the beginning of the \gls{aad} process, the code-feature guided mutation seems very beneficial. While both methods reach similar performance at the end of the budget, LLaMEA-SAGE achieves better performance faster and performance varies less across different runs. In Figure \ref{fig:SBOX-speedup} we clearly see that LLaMEA-SAGE is beneficial especially at the beginning of the runs, when looking at the relative speed-up of LLaMEA-SAGE versus LLaMEA.

To statistically verify the effect size and significance of our results, we calculated the area under the convergence curve (AUC) per seed for LLaMEA-SAGE and LLaMEA. On these per-seed AUC values, we applied a paired Wilcoxon signed-rank test and additionally report effect sizes (Cliff's $\delta$) and bootstrap confidence intervals for the mean AUC difference to account for the limited number of runs.
On SBOX-COST, LLaMEA-SAGE achieves a higher AUC than LLaMEA, with an average improvement of 11.1 AUC units. The effect size is large (Cliff's $\delta$ = 0.60), indicating a consistently faster convergence across runs. However, due to the limited number of seeds, the 95\% bootstrap confidence interval of the AUC difference remains wide and the paired Wilcoxon test does not reach statistical significance ($p = 0.44$).

\begin{figure}[!tb]
    \centering
    \includegraphics[width=1.0\linewidth,trim=3mm 4mm 3mm 3mm,clip]{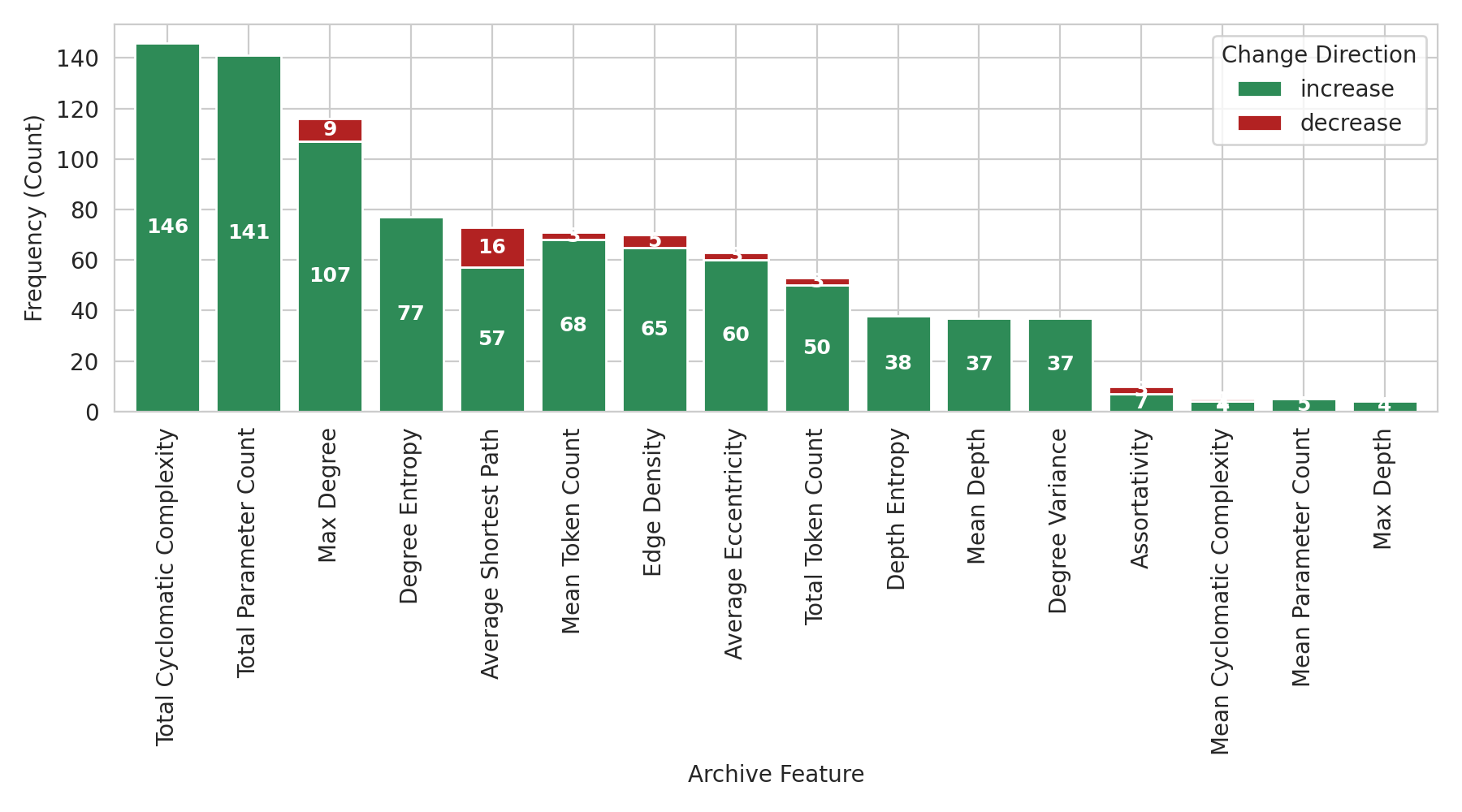}
    \caption{Code features and direction used during the $5$ different runs to guide the automated design of black-box optimization algorithms.}
    \label{fig:features-sbox}
\end{figure}

Next, we examine which code features have been used to steer the \gls{aad} process and with what suggested direction (increase versus decrease) in the $5$ runs. In Figure \ref{fig:features-sbox}, we can observe that a large variety of code features has been used to at least some extent, with \textit{total cyclomatic complexity} and \textit{total parameter count} being the most frequently used code-features. It is notable that the suggested direction is almost always ``increase'', indicating that increasing the complexity (one way or another) of the generated solutions is beneficial for the performance on this task. We emphasize that this trend is \textit{benchmark-specific} and should not be interpreted as a general preference for more complex algorithms.

\begin{figure}[!tb]
    \centering
    \includegraphics[width=0.48\linewidth,trim=0mm 4mm 3mm 3mm,clip]{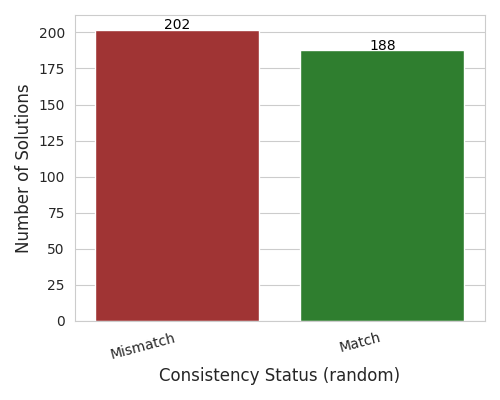}
    \includegraphics[width=0.48\linewidth,trim=0mm 4mm 3mm 3mm,clip]{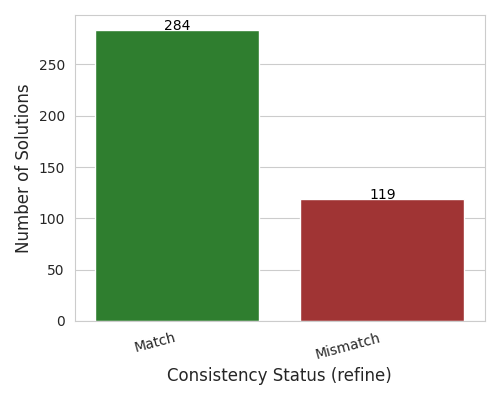}
    \caption{Consistency between code feature--guided suggestions and actual code feature mutation by LLaMEA-SAGE for different mutation prompts (left the \textbf{random new} and right the \textbf{refine} prompt).}
    \label{fig:consistency}
\end{figure}

To further investigate our newly proposed code feature--guidance method, we calculate the code feature difference between parent and offspring and how often this is consistent with the provided prompt guidance message. For example, if LLaMEA-SAGE asks the LLM to refine the given solution and increase the total token count, the resulting individual can either have a larger token count than its parent (Match) or a lower (Mismatch). The results of this analysis in Figure \ref{fig:consistency} tells a clear message: when the LLM is informed to generate a completely \textbf{new} algorithm, it ignores the code feature guidance completely, while when the \textbf{refine} prompt is used, it mostly follows orders (but not always).

Using \textit{Code Evolution Graphs} (CEG)~\cite{van2025code}, we can analyse in detail what happens during the different mutation steps. A CEG is a graph where we display each generated algorithm as a node and each parent-child relation in the \gls{aad} process as an edge between its parent and child node. The node color represents normalized fitness where yellow nodes are of high fitness and dark blue nodes of low fitness. For this paper, we have modified the standard CEG to also show information about what kind of mutation prompt has been used, by coloring the edges either yellow (\textbf{random new} prompt) or light blue (\textbf{refine} prompt). In addition, we use a dark green color when the prompt contains a guidance to \textit{increase} a given code feature together with the \textbf{refine} prompt and a dark blue color when the guidance contains an \textit{increase} together with the \textbf{random new} prompt.

\begin{figure}[!bt]
    \centering
    \includegraphics[width=1.\linewidth,trim=0mm 3mm 6mm 0mm,clip]{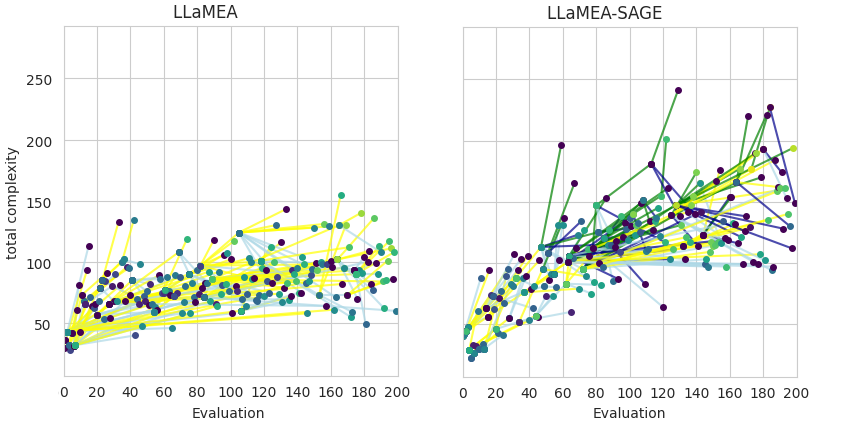}
    \caption{Code evolution graph of the total cyclomatic complexity of designed algorithms over time for one run of LLaMEA and LLaMEA-SAGE. Node colour represents normalized fitness where yellow nodes are of high fitness and dark blue nodes of low fitness. A yellow edge denotes the use of the \textbf{refine prompt}, light blue denotes the use of the \textbf{random new} prompt, green denotes an increase of the code feature and the use of the \textbf{refine} prompt, dark blue denotes an increase and the use of the \textbf{random new} prompt.}
    \label{fig:ceg-complexity-sbox}
\end{figure}

In Figure~\ref{fig:ceg-complexity-sbox}, we can see how the total cyclomatic complexity of generated algorithms evolves over the run using different mutation prompts for LLaMEA and LLaMEA-SAGE. We track the \texttt{total cyclomatic code complexity} as the selected feature guidance metric for the coloring mentioned above. From Figure~\ref{fig:ceg-complexity-sbox}, we can see that the code feature guidance does cause a larger exploratory effect in total complexity versus a non-guided mutation (vanilla LLaMEA). We can also clearly observe that the \textbf{random new} prompt almost always lead to lower complexity solutions, while the \textbf{refine} prompt is most often increasing the code complexity. Additional CEGs for different code features for all runs are provided in the supplementary material (Appendix C).

\subsection{Many Affine BBOB Competition}
\begin{figure}[!tb]
    \centering
    \includegraphics[width=1.0\linewidth,trim=3mm 4mm 4mm 9mm,clip]{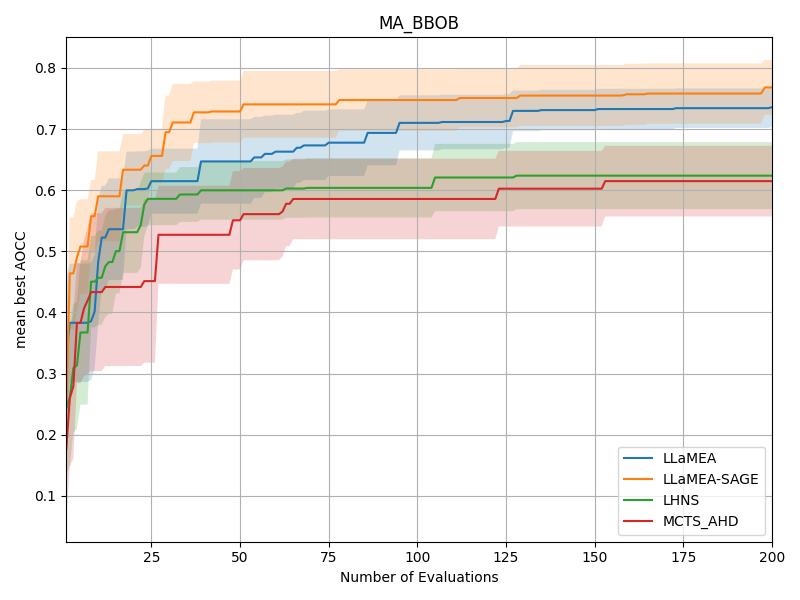}
    \caption{Average best-so-far fitness (AOCC) over time of the SOTA baselines and the proposed code feature guided approach (LLaMEA-SAGE) on the MA-BBOB suite. Averaged over $5$ independent runs.}
    \label{fig:MABBOB}
\end{figure}

\begin{figure}[!tb]
    \centering
    \includegraphics[width=1.0\linewidth,trim=3mm 0mm 4mm 0mm,clip]{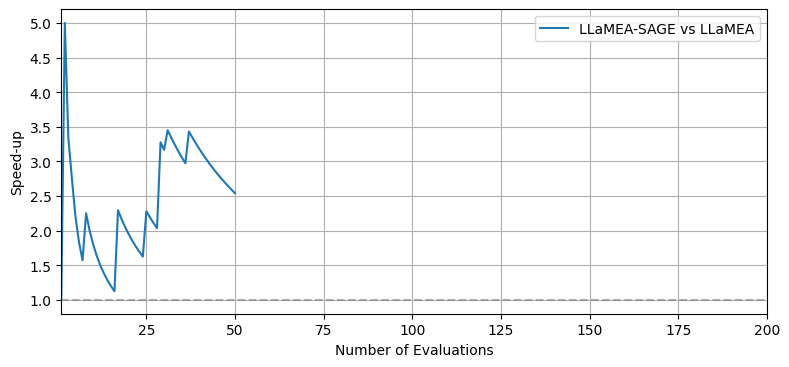}
    \caption{Average speed-up of LLaMEA-SAGE versus vanilla LLaMEA on MA-BBOB. The speed-up is how much faster than LLaMEA the proposed method is, i.e.\ if LLaMEA-SAGE reaches fitness value $f_i$ in $m$ evaluations and LLaMEA reaches that same fitness value in $n$ evaluations, the speed-up is $n/m$. Speed-up values after 50 evaluations are missing because LLaMEA never reaches the fitness value of LLaMEA-SAGE.}
    \label{fig:MABBOB-speedup}
\end{figure}

Now that we better understand how our proposed mutation guidance technique influences the \gls{aad} process, we perform a larger benchmarking exercise on the MA-BBOB suite. This experiment aims to validate the proposed method \textit{against strong state-of-the-art baselines}.

\begin{figure*}[!ht]
    \centering
    \includegraphics[width=0.95\linewidth,trim=0mm 3mm 3mm 0mm,clip]{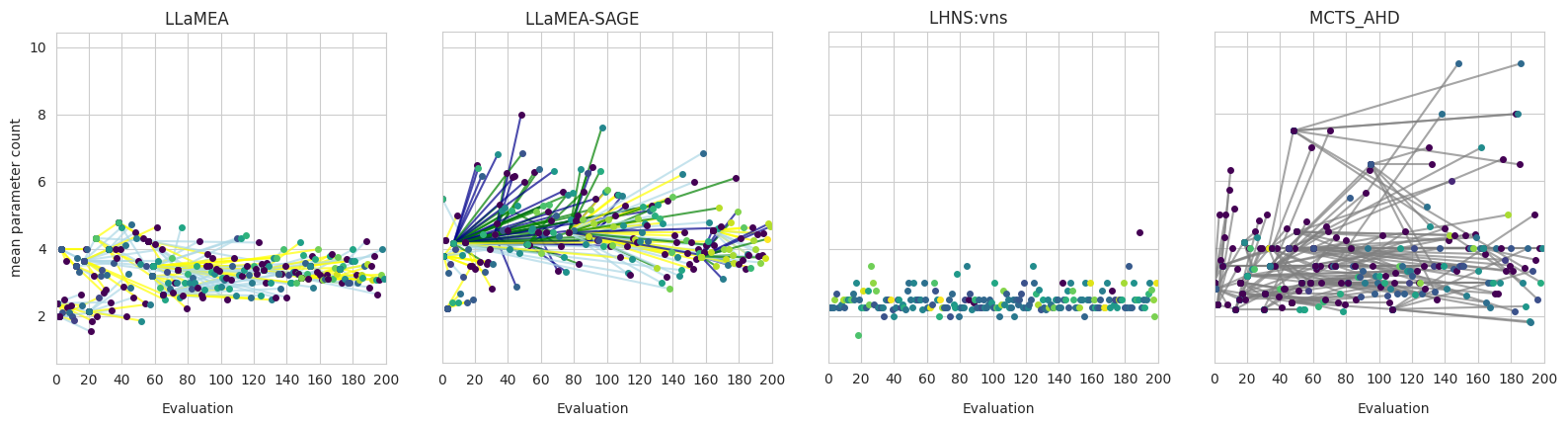}
    \caption{Code evolution graph of the mean parameter count of designed algorithms over time for the different methods on MA-BBOB. Colors of nodes and edges are according to Figure \ref{fig:ceg-complexity-sbox}}
    \label{fig:ceg-param-mabbob}
\end{figure*}

In Figure \ref{fig:MABBOB}, we can see the aggregated fitness over time for the proposed LLaMEA-SAGE and SOTA baselines. Again, we see a performance difference, especially in the beginning of the evolutionary process, between vanilla LLaMEA and LLaMEA-SAGE as can also be observed by the speed-up shown in Figure \ref{fig:MABBOB-speedup}. There we can also see that at only $50$ evaluations, LLaMEA-SAGE reaches an average fitness that LLaMEA cannot achieve even after the full $200$ evaluations. Both LLaMEA-based algorithms clearly outperform the LHNS and MCTS-AHD methods on this black-box optimization algorithm design task.

Also for MA-BBOB, statistical testing reveals that LLaMEA-SAGE achieves a higher mean AUC than vanilla LLaMEA but this time the effect size is medium (Cliff's $\delta = 0.36$), indicating a consistent but modest improvement. Due to the small number of runs (5 seeds), the 95\% bootstrap confidence interval of the AUC difference is wide and includes zero and the paired Wilcoxon test does not reach statistical significance (p = 0.44). Due to the substantial computational and API costs associated with large language model–based optimization, the number of runs was necessarily limited.

Figure~\ref{fig:ceg-param-mabbob} shows the CEG for each method, analysing the \texttt{mean parameter count} code feature on the $y$-axis. Note that for LHNS there is no parent-child relation logged and such there are no edges in the graph for that method. We observe is that for this particular code feature again there is a lot of exploration going on in LLaMEA-SAGE, especially when this feature is mentioned in the guidance prompt (dark blue and dark green colors). For LHNS, the value of the feature is relatively stable and low, while for MCTS-AHD, there are very large jumps in its feature value.

\subsection{Token Costs per Method}
\begin{figure}[!tb]
    \centering
    \includegraphics[width=0.7\linewidth,trim=2mm 2mm 2mm 2mm,clip]{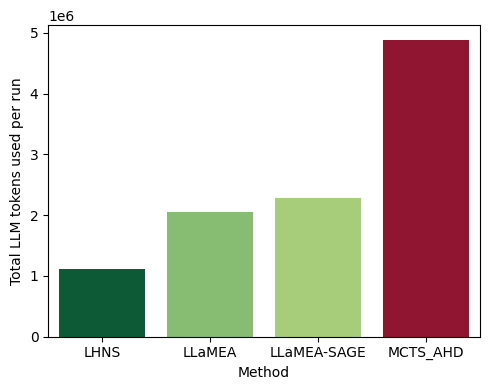}
    \caption{Average number of LLM tokens used per AAD method for one run of generating $200$ algorithms.}
    \label{fig:tokencost}
\end{figure}

As the usage of LLMs can be expensive, either in compute or in API costs, we also compare the number of LLM tokens required to do a full AAD run per method. In Figure \ref{fig:tokencost}, the average total number of tokens per run is shown. We can observe that LHNS is the most token efficient method, this is most likely not because of greater efficiency of the approach, but because it tends to generate less complex code, which consumes fewer tokens. LLaMEA and LLaMEA-SAGE are very close to each other, with the additional code feature guidance increasing the token cost only marginally. This shows that LLaMEA-SAGE \textit{preserves scalability} relative to vanilla LLaMEA. MCTS-AHD has by far the highest token cost (more than twice as high as LLaMEA) as it queries the LLM several times per solution, while the other methods only query the LLM once per generated algorithm.

\subsection{Validation of Final Best Algorithms}
\begin{figure*}[!tb]
    \centering
    \includegraphics[width=0.32\linewidth,trim=0mm 4mm 3mm 2mm,clip]{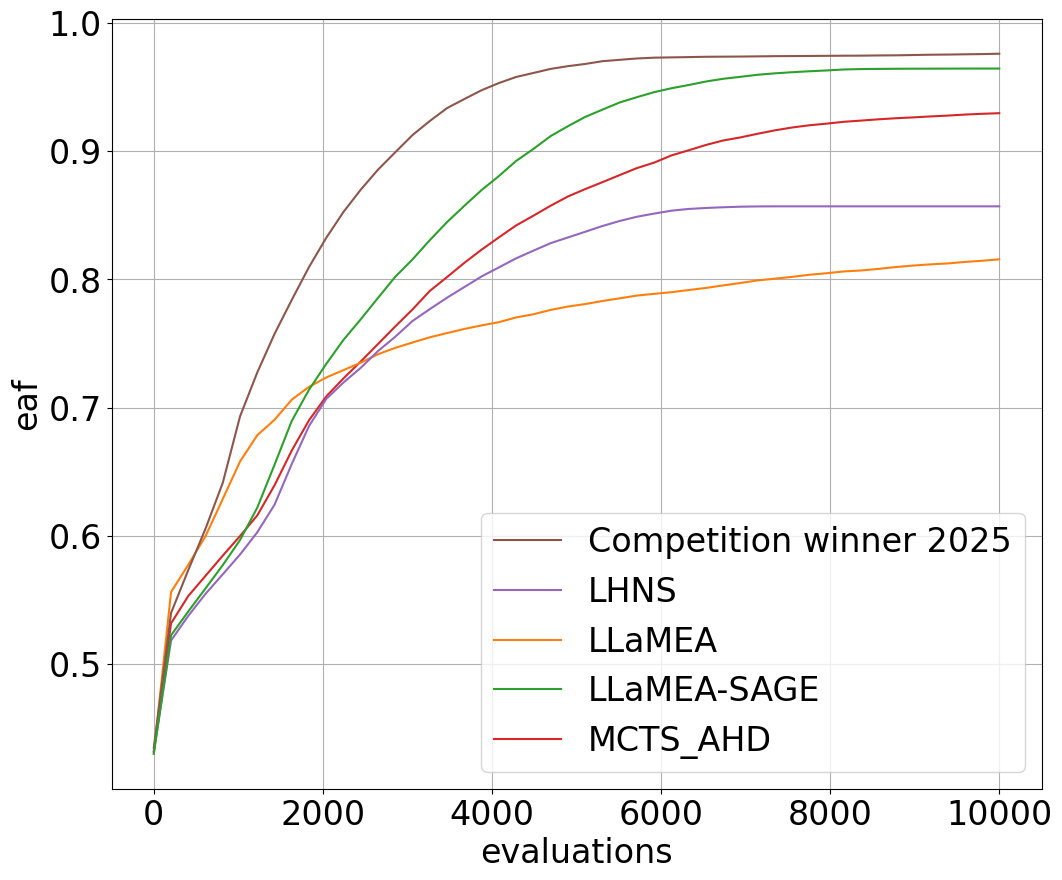}
    \includegraphics[width=0.32\linewidth,trim=0mm 4mm 3mm 2mm,clip]{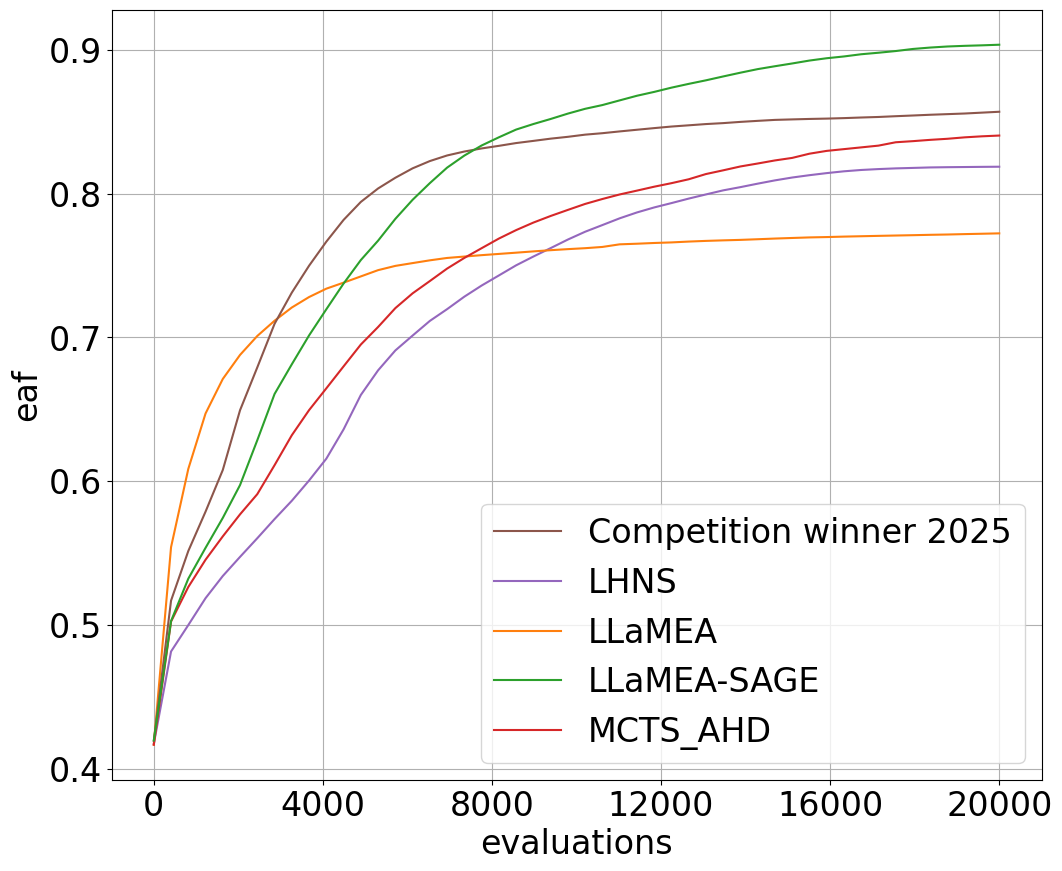}
    \includegraphics[width=0.32\linewidth,trim=0mm 4mm 3mm 2mm,clip]{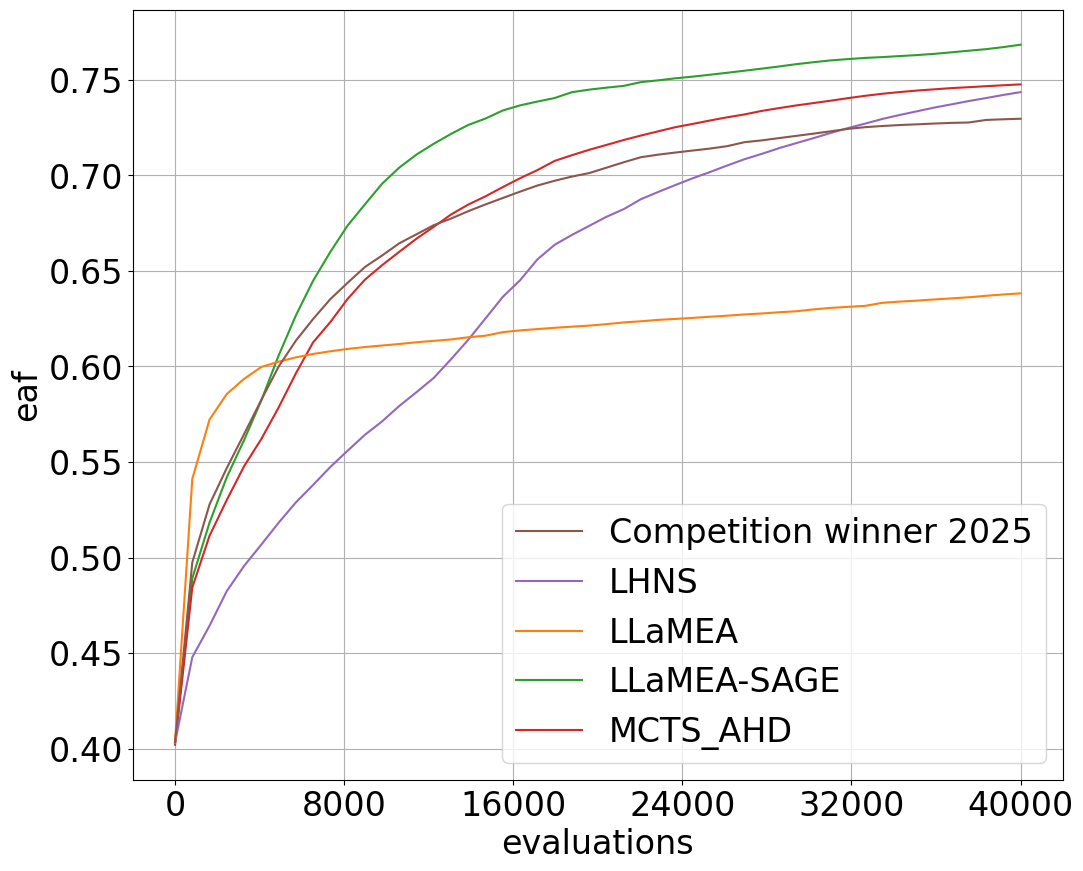}
    \caption{Empirical attainment function based ECDF curves (higher is better) on $50$ MA-BBOB instances in $5d$ (left plot), $10d$ (middle plot) and $20d$ (right plot) for all best discovered algorithms during the runs of different AAD methods. All algorithms have been trained only on $10d$ problem instances apart from the competition winner, which has been trained on $5d$.}
    \label{fig:validation}
\end{figure*}

We validate the final best generated algorithm from each approach, i.e.\ the generated algorithm with the highest fitness over the 5 independent runs per method (LLaMEA, LLaMEA-SAGE, LHNS and MCTS). For the final evaluation we use $50$ instances from MA-BBOB (instances $100$ to $150$) and $d=5$, $d=10$ and $d=20$ with a budget of $2000 \cdot d$ function evaluations. Two out of the three comparisons are performed on dimensionalities different from the one used for the AAD to assess how well the generated algorithms generalise. We also include in the comparison the winner of the 2025 MA-BBOB competition (NeighborhoodAdaptiveDE)~\cite{van2025neighborhood} (specifically tuned on MA-BBOB $5d$).

The empirical attainment function (EAF) based empirical cumulative distribution function (ECDF) curves from the three validation experiments can be seen in Figure \ref{fig:validation}.
The EAF estimates the percentage of runs that attain a given target value no later than a given runtime. Taking the partial integral of the EAF results in a more accurate version of the Empirical Cumulative Distribution Function, as it does \textit{not} rely on discretization of the targets. Larger EAF values means that the algorithm performs better.
While the best generated algorithm from the approach we propose here does not beat the competition winner in $5d$ (on which NeighborHoodAdaptiveDE was tuned), it does $10d$ and $20d$ and still performs well on $5d$, in particular beating all other automatically designed algorithms.

\section{Conclusions}
\label{sec:conclusions}

We introduced \emph{LLaMEA-SAGE}, a structured feedback mechanism for automated algorithm design that augments LLM-driven evolutionary search with information extracted from program code. By analysing abstract syntax tree (AST) and complexity code features of generated algorithms, learning a surrogate model over an archive of evaluated solutions and using explainable AI techniques to identify influential structural properties, we close the loop between code structure and black-box optimization performance. Crucially, this feedback is translated into lightweight natural-language guidance prompts that steers LLM-based mutations without constraining expressivity or imposing rigid templates.

Our experimental results demonstrate that code feature guided mutation improves the efficiency and stability of automated algorithm discovery. In controlled experiments, our proposed LLaMEA-SAGE achieves good performance faster than vanilla LLaMEA under identical budgets, while extensive analysis using code evolution graphs shows that the guidance meaningfully biases structural exploration. In a large-scale comparison on the MA-BBOB benchmark suite, LLaMEA-SAGE consistently outperforms strong state-of-the-art baselines, including MCTS-based and large-neighborhood search approaches, while retaining the flexibility of template-free, end-to-end algorithm generation. These findings suggest that even relatively simple structural signals can provide a powerful inductive bias for LLM-driven algorithm evolution.

\paragraph{Limitations and future work.}
While the results are promising, several limitations remain. First, our experiments are conducted using a single LLM backend, while ablation studies (see the Supplementary material Appendix B) on a different LLM backend (Gemini-2.0-flash-lite) show a consistent outcome; a more systematic study across more models would be beneficial to assess robustness. Second, the current evaluation focuses on continuous black-box optimization benchmarks at moderate dimensionalities; it remains an open question how well the proposed guidance generalizes to higher-dimensional settings, noisy objectives or fundamentally different problem classes. Third, the code feature set, while deliberately generic, captures only static structural properties of code and does not account for dynamic execution behavior or interaction effects during optimization runs.

These limitations point to several directions for future research. Promising extensions include combining structural code features with runtime or behavioral descriptors, exploring adaptive or multi-feature guidance strategies instead of single-feature attribution and integrating the proposed mechanism into other LLM-based algorithm design frameworks beyond LLaMEA. More broadly, we believe that closing the loop between program analysis, explainable models and generative LLMs is a key step toward more principled, interpretable and scalable automated algorithm design.

\FloatBarrier


\bibliographystyle{ACM-Reference-Format}
\bibliography{bibliography}

\appendix

\onecolumn

\section*{\Huge Supplementary Material for LLaMEA-SAGE}

\section{Problem Specific Prompts used}
\label{sec:appendix-prompts}

This appendix provides the exact problem specification prompts used to initialize the AAD process in our experiments. Including these prompts verbatim is essential for reproducibility, as they define the optimization task, interface constraints and permissible implementation details exposed to the LLM.

Appendix~A.1 documents the prompt used for the SBOX-COST benchmark in Experiment~1, while Appendix~A.2 reports the corresponding prompt for the MA-BBOB benchmark used in Experiment~2. 

All AAD methods compared in this paper (LLaMEA, LLaMEA-SAGE, LHNS, and MCTS-AHD) were initialized using the same problem specification prompts for a given benchmark. Differences in performance therefore arise from the search strategies and mutation mechanisms rather than differences in task formulation.

\subsection{SBOX}

\begin{subbox}{Problem specification prompt}
\Tiny
\begin{verbatim}
You are an excellent Python programmer.
You are a Python expert working on a new optimization algorithm. 
You can use numpy v2 and some other standard libraries.
Your task is to develop a novel heuristic optimization algorithm for continuous optimization problems. 
The optimization algorithm should work on different instances of noiseless unconstrained functions. 
Your task is to write the optimization algorithm in Python code. 
Each of the optimization functions has a search space between -5.0 (lower bound) and 5.0 (upper bound). 
The dimensionality can be varied.
The code should contain an `__init__(self, budget, dim)` function with optional additional arguments 
and the function `def __call__(self, func)`, 
which should optimize the black box function `func` using `self.budget` function evaluations.
The func() can only be called as many times as the budget allows, not more. 

An example of such code (a simple random search), is as follows:
```python
import numpy as np
import math

class RandomSearch:
    def __init__(self, budget=10000, dim=10):
        self.budget = budget
        self.dim = dim
    def __call__(self, func):
        self.f_opt = np.inf
        self.x_opt = None
        for i in range(self.budget):
            x = np.random.uniform(func.bounds.lb, func.bounds.ub)
            
            f = func(x)
            if f < self.f_opt:
                self.f_opt = f
                self.x_opt = x
            
        return self.f_opt, self.x_opt
```
Give an excellent and novel heuristic algorithm to solve this task and also give it a one-line description, 
describing the main idea. Give the response in the format:
# Description: <short-description>
# Code: 
```python
<code>
```
\end{verbatim}
\end{subbox}

\subsection{MA-BBOB}

\begin{subbox}{Problem specification prompt}
\Tiny
\begin{verbatim}
You are an excellent Python programmer.
You are a Python developer working on a new optimization algorithm.
Your task is to develop a novel heuristic optimization algorithm for continuous optimization problems.
The optimization algorithm should handle a wide range of noiseless functions. 
Your task is to write the optimization algorithm in Python code. 
Each of the optimization functions has a search space between -5.0 (lower bound) and 5.0 (upper bound). 
The dimensionality can be varied.
The code should contain an `__init__(self, budget, dim)` function with optional additional arguments 
and the function `def __call__(self, func)`, which should optimize the black box function `func` using `self.budget` function evaluations.
The func() can only be called as many times as the budget allows, not more. 

An example of such code (a simple random search), is as follows:
```python
import numpy as np

class RandomSearch:
    def __init__(self, budget=10000, dim=10):
        self.budget = budget
        self.dim = dim

    def __call__(self, func):
        self.f_opt = np.inf
        self.x_opt = None
        for i in range(self.budget):
            x = np.random.uniform(func.bounds.lb, func.bounds.ub)
            
            f = func(x)
            if f < self.f_opt:
                self.f_opt = f
                self.x_opt = x
            
        return self.f_opt, self.x_opt
```

Give an excellent and novel heuristic algorithm to solve this task and also give it a one-line description, 
describing the main idea. Give the response in the format:
# Description: <short-description>
# Code: 
```python
<code>
```
", "cost": 0.0001131, "tokens": 0}

\end{verbatim}

\end{subbox}

\section{LLM Ablation studies}
\label{sec:appendix-llm}

This appendix reports an additional ablation experiment assessing the robustness of the proposed code feature guided mutation mechanism on a different LLM backend. While the main paper focuses on results obtained using the GPT-5-mini model, we repeat experiment 2 using the Gemini-2.0-flash-lite model to evaluate whether the observed performance trends are model-specific.

\begin{figure}[H]
    \centering
    \includegraphics[width=.48\linewidth]{figures/MABBOB-aocc.png}
    \includegraphics[width=.48\linewidth]{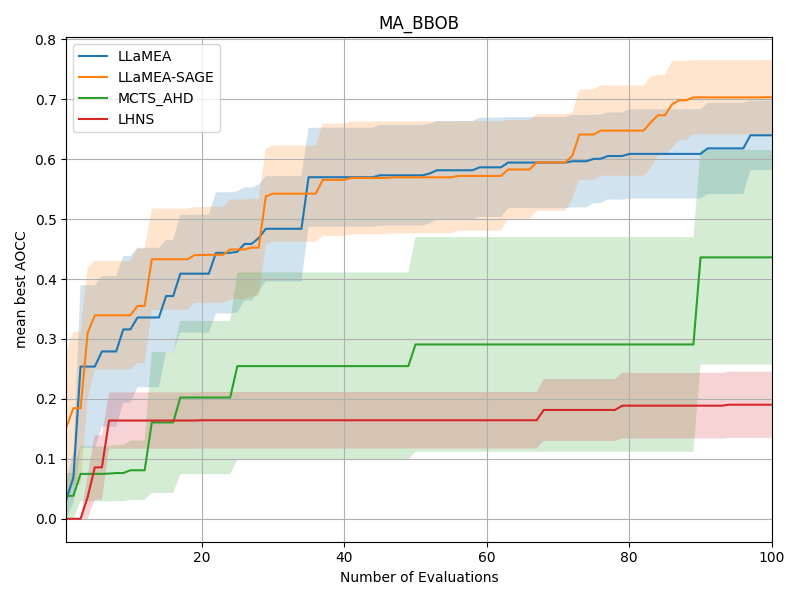}
    \caption{Average best-so-far fitness (AOCC) over time of the SOTA baselines and the proposed code feature guided approach (LLaMEA-SAGE) on the MA-BBOB suite using on the left \textbf{gpt-5-nano} vs on the right \textbf{gemini-flash-2.0-lite}. Averaged over $5$ independent runs.}
    \label{fig:MABBOB-gemini}
\end{figure}

Figure \ref{fig:MABBOB-gemini} compares the anytime performance (AOCC) of LLaMEA-SAGE and baseline methods on the MA-BBOB benchmark under both LLM backends. Although absolute performance levels differ, the relative advantage of code feature guided mutation remains consistent.

These results support the claim that the proposed guidance mechanism captures structural signals that generalize across LLM implementations, rather than exploiting idiosyncrasies of a single model.

\section{Additional Code Evolution Graphs}

\label{sec:appendix-ceg}

This appendix provides additional Code Evolution Graphs (CEGs) to complement the analyses presented in Section~5. While the main paper highlights representative runs and selected code features, the figures in this appendix offer a more comprehensive view across benchmarks, runs and structural properties.

First we show CEGs for the SBOX-COST benchmark, including the evolution of features such as mean parameter count, maximum degree and token count over time. Next we report the corresponding graphs for the MA-BBOB benchmark.

Across these figures, node color encodes normalized fitness, while edge color indicates the mutation prompt type and whether code feature guidance requested an increase or decrease of the selected feature (only for LLaMEA-SAGE). Together, these visualizations illustrate how code feature guidance systematically biases structural exploration during the AAD process, while preserving substantial diversity in the generated algorithms.

\begin{figure*}[p]
    \centering
    \includegraphics[width=1.0\linewidth]{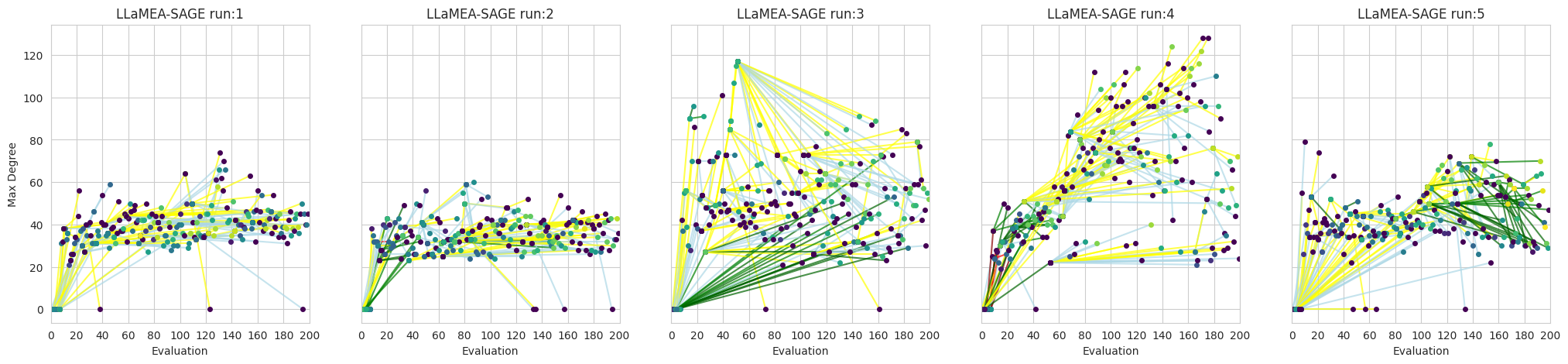}
    \caption{Code evolution graph of the max degree code feature of designed algorithms over time for SBOX-COST. Node colour represents normalized fitness where yellow nodes are of high fitness and dark blue nodes of relative low fitness. A yellow edge denotes the use of the \textbf{refine prompt}, light blue denotes the use of the \textbf{random new} prompt, green denotes an increase of the code feature and the use of the \textbf{refine} prompt, dark blue denotes an increase and the use of the \textbf{random new} prompt}
    \label{fig:max-degree-sbox}
\end{figure*}

\begin{figure*}[p]
    \centering
    \includegraphics[width=1.0\linewidth]{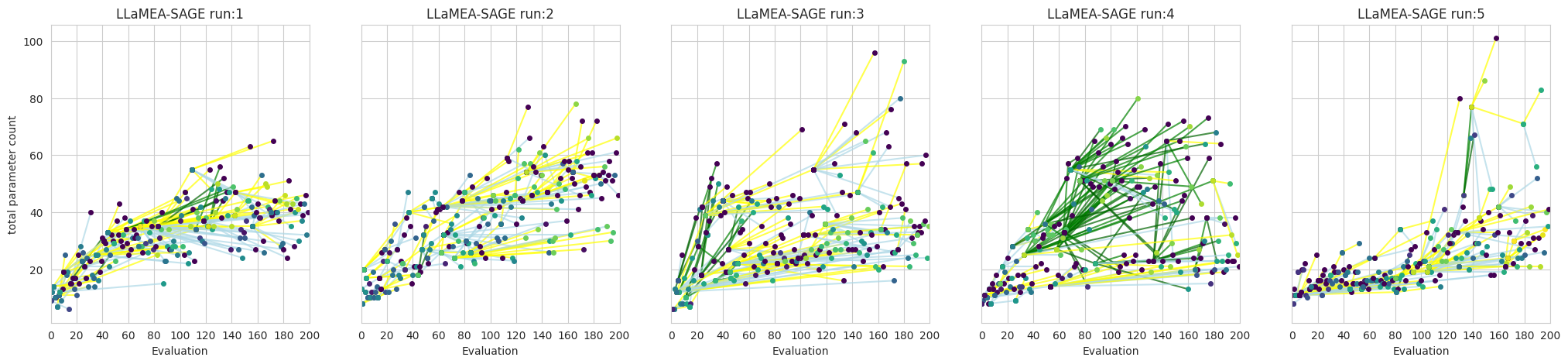}
    \caption{Code evolution graph of the mean parameter count code feature of designed algorithms over time for SBOX-COST. Node colour represents normalized fitness where yellow nodes are of high fitness and dark blue nodes of relative low fitness. A yellow edge denotes the use of the \textbf{refine prompt}, light blue denotes the use of the \textbf{random new} prompt, green denotes an increase of the code feature and the use of the \textbf{refine} prompt, dark blue denotes an increase and the use of the \textbf{random new} prompt}
    \label{fig:param-count-sbox}
\end{figure*}

\begin{figure*}[p]
    \centering
    \includegraphics[width=1.0\linewidth]{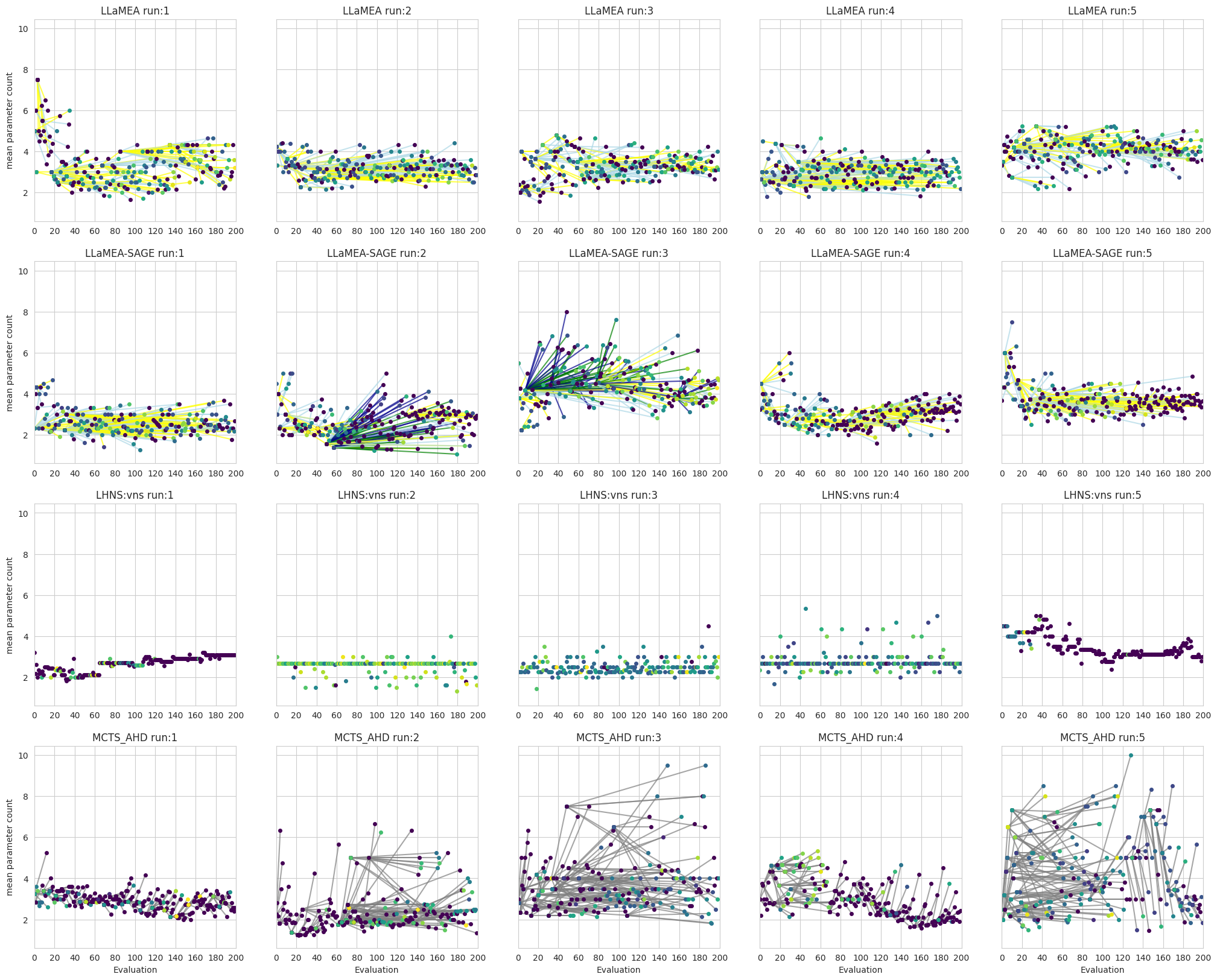}
    \caption{Code evolution graph of the mean parameter count code feature of designed algorithms over time for MA-BBOB. Node colour represents normalized fitness where yellow nodes are of high fitness and dark blue nodes of relative low fitness. A yellow edge denotes the use of the \textbf{refine prompt}, light blue denotes the use of the \textbf{random new} prompt, green denotes an increase of the code feature and the use of the \textbf{refine} prompt, dark blue denotes an increase and the use of the \textbf{random new} prompt}
    \label{fig:param-count-mabbob}
\end{figure*}

\begin{figure*}[p]
    \centering
    \includegraphics[width=1.0\linewidth]{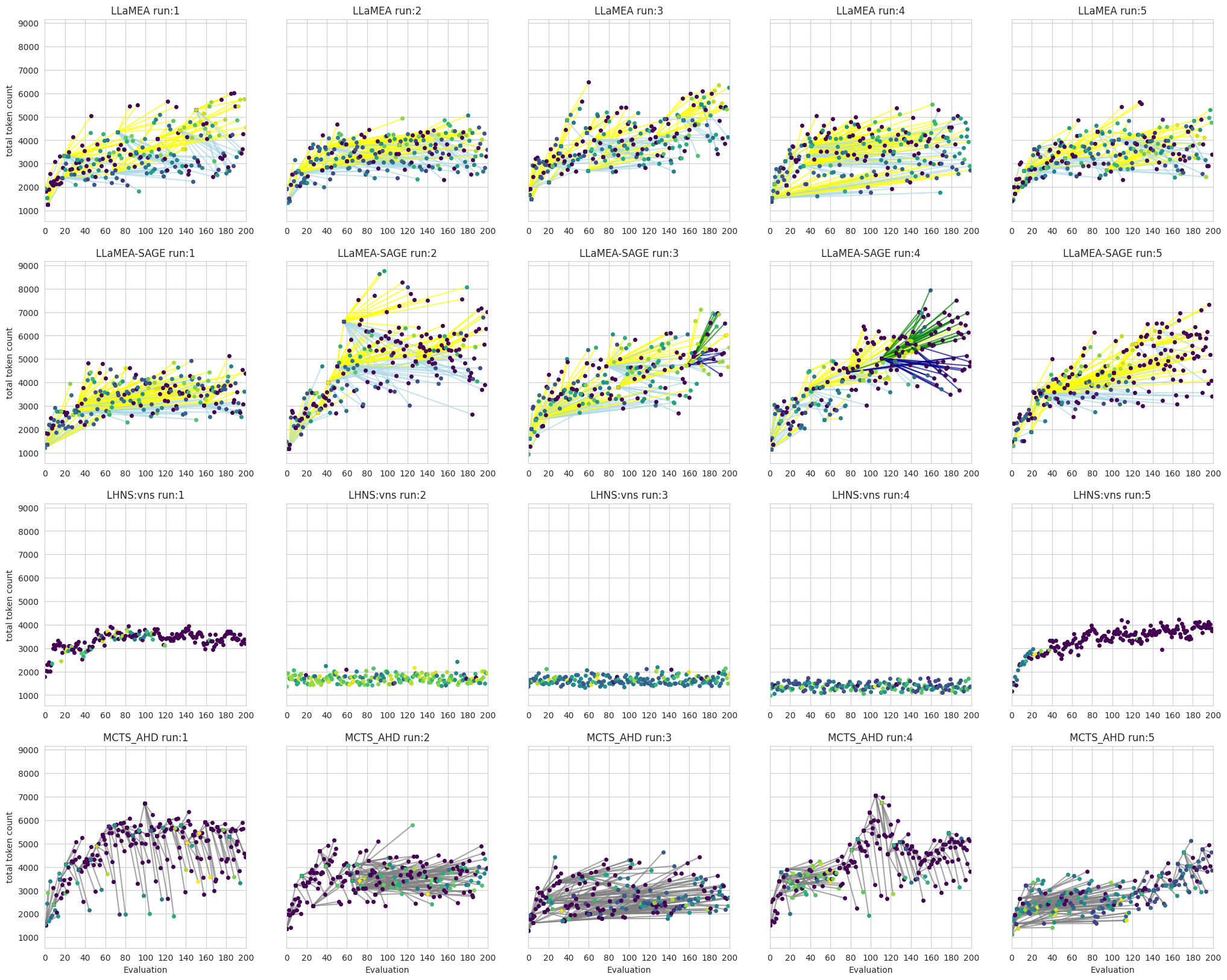}
    \caption{Code evolution graph of the total number of code tokens  of designed algorithms over time for MA-BBOB. Node colour represents normalized fitness where yellow nodes are of high fitness and dark blue nodes of relative low fitness. A yellow edge denotes the use of the \textbf{refine prompt}, light blue denotes the use of the \textbf{random new} prompt, green denotes an increase of the code feature and the use of the \textbf{refine} prompt, dark blue denotes an increase and the use of the \textbf{random new} prompt}
    \label{fig:token-count-mabbob}
\end{figure*}

\end{document}